\documentclass{article}
\usepackage{spconf,amsmath,graphicx}
\usepackage{hyperref}
\usepackage[sort, numbers]{natbib}
\usepackage{multirow}
\usepackage{enumitem}
\title{Robust Skin Color Driven Privacy-Preserving Face Recognition via Function Secret Sharing}
\name{
    Dong Han$^{\star \dagger}$ \qquad
    Yufan Jiang$^{\star}$ \qquad
    Yong Li$^{\star \ddagger}$\thanks{$\ddagger$ Corresponding Author} \qquad
    Ricardo Mendes$^{\star}$ \qquad
    Joachim Denzler$^{\dagger}$
}
\address{
    $^{\star}$Huawei Munich Research Center, Germany \\
    $^{\dagger}$Computer Vision Group, Friedrich Schiller University Jena, Germany
}
\begin{document}
%\ninept
%
\maketitle
%
% \renewcommand{\thefootnote}{\fnsymbol{footnote}}
% \footnotetext[3]{Corresponding author}
% \footnote{$^{\ddagger}$Corresponding author}

\begin{abstract}

In this work, we leverage the pure skin color patch from the face image as the additional information to train an auxiliary skin color feature extractor and face recognition model in parallel to improve performance of state-of-the-art (SOTA) privacy-preserving face recognition (PPFR) systems. Our solution is robust against black-box attacking and well-established generative adversarial network (GAN) based image restoration. We analyze the potential risk in previous work, where the proposed cosine similarity computation might directly leak the protected precomputed embedding stored on the server side. We propose a Function Secret Sharing (FSS) based face embedding comparison protocol without any intermediate result leakage. In addition, we show in experiments that the proposed protocol is more efficient compared to the Secret Sharing (SS) based protocol.

\end{abstract}

\section{Introduction}\label{sec:intro}

\begin{figure*}[t!]
\centering
\includegraphics[scale=0.48]{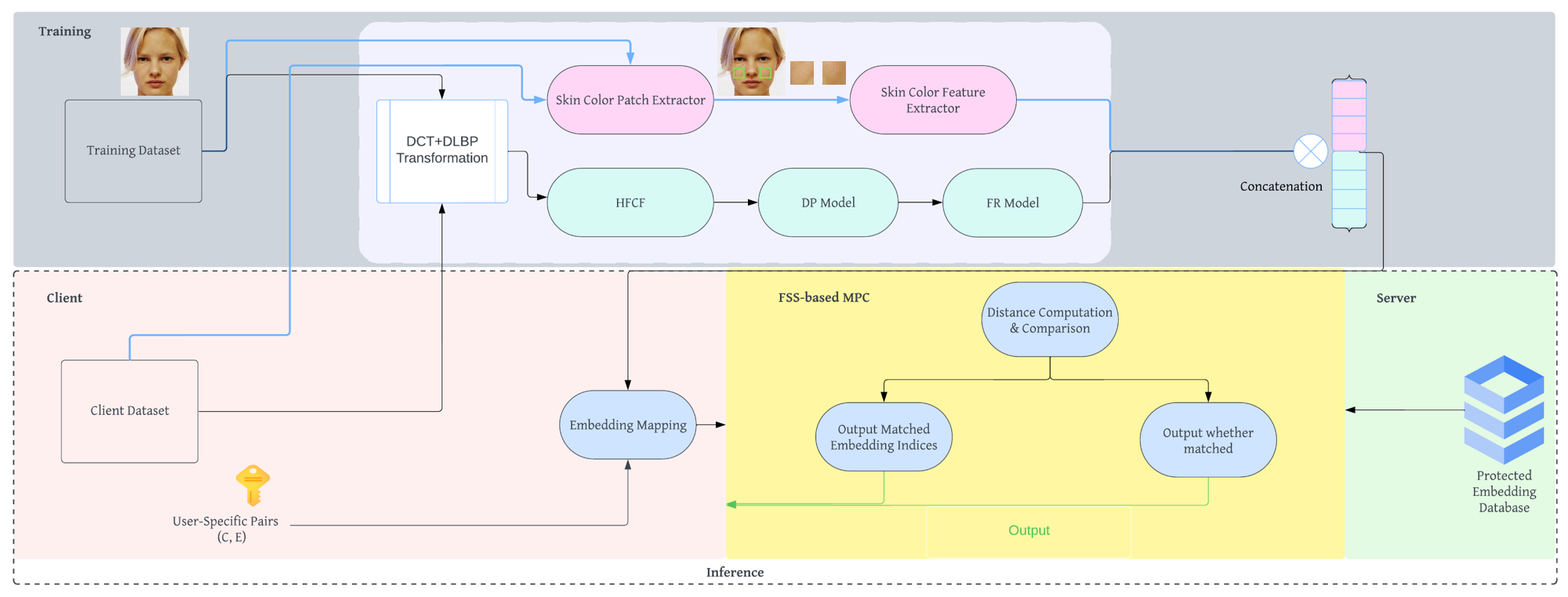}
\caption{Overview of proposed HFCF-SkinColor privacy-preserving FR framework with FSS-based Distance Computation.}
\label{fig:workflow}
\end{figure*}

With the recent advance of artificial intelligence (AI), the utility of AI technology is expanding to various domains. 
However, misuse, privacy invasion, and security concerns are sprouting up. Hence, trustworthy AI aims to reduce the potential risks from original data, models and downstream applications while promising reliability, robustness and explainability. 
Face recognition (FR) is one of the most crucial applications.

The privacy-preserving face recognition (PPFR) techniques can be classified into two types based on whether the privacy-preserving technique is applied to input face images or to face embeddings. For the former, the goal is to introduce artifacts into the original face image or transfer the image to another sparse representation with certain visual information concealed. For the latter, the algorithm is targeted at the face embedding protection which encrypts or maps the original vector into protected space. The homomorphic encryption based PPFR method trains face images in ciphertext form and encrypts embeddings for face matching \cite{yang2022design}.
% Differential privacy (DP) is another common method to protect data privacy by adding perturbation to face images \cite{chamikara2020privacy}. However, it tends to considerably degrade the accuracy of face recognition. 
FaceMAE \cite{wang2022facemae} takes advantage of the image completion ability of the masked autoencoders (MAE) to reconstruct the approximation of a randomized masked face. However, there is a potential privacy leakage since the backbone model still needs to ``see'' the reconstructed face image. Instead of adding artifacts or manipulating face images directly in RGB color space, PPFR-FD \cite{wang2022privacy} argues convolutional neural networks (CNNs) can be trained with only high-frequency information in Discrete Cosine Transform (DCT) representation for FR. 
%The controllable privacy budget for PPFR is proposed by adding differential privacy (DP) noise to DCT images \cite{ji2022privacy}. 
Differential privacy (DP), the de facto privacy standard, is also proposed as a PPFR, as a way to add controllable noise to DCT images \cite{ji2022privacy}. 
Another work \cite{han2024privacypreserving} incorporates frequency and color information together to improve recognition accuracy. The face embedding comprises vectorized soft-biometric properties associated with each identity. It is possible to map back to a 2D face image only based on such a 1D embedding vector \cite{vendrow2021realistic}. Thus, it is necessary to protect not only the input face images but also the corresponding embeddings. The PolyProtect \cite{hahn2022towards} is an irreversible method to transfer the embedding into a low-dimensional protected space with user-specific parameters. Multi-IVE \cite{melzi2023multi} is based on Incremental Variable Elimination (IVE) to protect the soft-biometric information of embeddings in the Principal Component Analysis (PCA) domain.

In contrast to the aforementioned frequency-based approach, we exploit the skin color information as an auxiliary feature to improve recognition accuracy while maintaining the privacy of face images. Besides, we propose a Function Secret Sharing (FSS) based approach to securely compute and compare the cosine similarity of protected embedding pairs with the threshold value, without leaking any intermediate results.
To summarize, the main contributions of this work are as follows:

\begin{itemize}[leftmargin=*]
  \item We introduce HFCF-SkinColor, a simple method for increasing recognition accuracy of PPFR in both 1:1 and 1:N verification situations.
  \item We show the robustness and trustworthiness of HFCF-SkinColor by experimenting black-box attacking.
  \item To accelerate the online computation, we propose an FFS-based face embedding distance comparison protocol for two different application scenarios without leaking any intermediate results.
\end{itemize}

% \begin{figure*}[t!]
% \centering
% \includegraphics[scale=0.61]{Images/workflow_skincolor_new.png}
% \caption{Overview of proposed HFCF-SkinColor privacy-preserving FR framework with FSS-based Distance Computation.}
% \label{fig:workflow}
% \end{figure*}

\section{PROPOSED METHOD}
\label{sec:format}

% There are several considerations when designing a HFCF-SkinColor alternative for the PPFR. We will address them in following sections. The proposed framework is shown in Figure \ref{fig:workflow}.

\subsection{Skin Color Information for PPFR}

The previous SOTA frequency-based FR methods transfer face image into frequency domain. Normally, the lowest frequency channel (direct current (DC)) is discarded since it contains the most visual information. However, the number of selected channels affects the recognition accuracy and computation time. Moreover, the color information is not utilized in those frequency-based models. Recent work \cite{han2024privacypreserving} fuses the frequency and color information by Hybrid Frequency-Color Fusion (HFCF) where the color information is represented as the decoded local binary pattern (DLBP) feature. % Authors in \cite{guo2023faceskin} declare 
Previous research \cite{guo2023faceskin} shows that the CNN-based classification model cannot converge with pure skin color patches to predict neither age nor race but color patches can improve accuracy when served as extra information along with face images for soft-biometric attribute classification. 
Inspired by this work, we employ pure skin patches as the additional information for improving PPFR performance, which is more challenging than predicting only the soft-biometric information. Another difficulty is that we use relatively low-resolution face images \cite{yao2024novel} compared with theirs, which can impede the performance of face landmark detection and skin patch localization.
As shown in Figure \ref{fig:workflow}, the proposed PPFR framework integrates a skin color extractor with HFCF to generate combined face embeddings.
During the training stage, original face information is transferred into a noised HFCF representation and skin color feautre for privacy preservation.
The client generates C, E pair for each identity for a secure query with protected embedding held by the server after embedding mapping.
% The potential alternative solution to improve the face detection is \cite{yao2024novel}.

\subsection{FFS-based Face Embedding Distance Comparison}

In the traditional FR system, the similarity between two query face images is determined according to the distance that calculated from corresponding embeddings. Here we introduce a novel embedding distance comparison protocol based on the function secret sharing scheme \cite{boyle2021function} in the \textit{preprocessing model}.
%\cite{boyle2015function,boyle2016function,boyle2021function,boyle2019secure}

While the secret sharing (SS) scheme shares the secret value itself, FSS shares the computed function to function keys.
To compute any FSS gate (e.g. a dReLU gate), parties input a masked value consistently, and output the masked shared computation result.
While the entire FSS gate computation remains local, the only necessary communication is to reveal the masked FSS gate wire inputs.
Thus, complex computation gates can be securely evaluated in one single round, which significantly improves the online stage of Multiparty Computation (MPC).
In this work, we mainly use the FSS-based dReLU protocol (comparison protocol) and the division protocol to implement the face embedding distance comparison.
%For more details, we refer the reader to the original work \cite{boyle2015function,boyle2016function,boyle2021function,boyle2019secure}.

% \section{Secure Embedding Comparison Via FSS}

\subsection{How the Security in HFCF-DLBP Fails}
HFCF-DLBP \cite{han2024privacypreserving} has proposed an MPC-based scheme to protect the precomputed embedding on the server side.
However, the security breaks if a client is allowed to make multiple comparison request to the server.
Suppose the client is corrupted by a \textbf{semi-honest} adversary, who does not violate the protocol but tries to infer private information from the server.
We note the precomputed embeddings on the server side as $\mathcal{P} = \{P_1, P_2,...,P_m\}$, where $P_i = \{P_{i,1},...,P_{i,n}\}$.
And we note the protected embedding on the client side as $P_c$.
We use $|\cdot|$ to denote the length of an embedding, typically we have $|P_i| = |P_c| = n$.
During each protocol execution round, parties reveal every dot product to the client, we note it as $D = \{d_1,...,d_m\}$.
If the client queries $n$ times, it receives $D^j = \{d^j_1,...,d^j_m\}$ for every query $j$.
Thus, the client is able to solve the following linear algebra for $1 \leq i \leq n$:
\[
\begin{bmatrix}
P^1_{c,1} & \dots & P^1_{c,n}\\
\vdots & \ddots & \vdots \\
P^n_{c,1} & \dots & P^n_{c,n}
\end{bmatrix}
\cdot
\begin{bmatrix}
P_{i,1} \\ 
\vdots \\
P_{i,n}
\end{bmatrix}
= 
\begin{bmatrix}
d^1_i \\
\vdots \\
d^n_i
\end{bmatrix},
\]
where $P_i$ is supposed to remain hidden.

\subsection{An FSS-based Solution}

\begin{figure}[h!]
    \centering
    \includegraphics[scale=0.49]{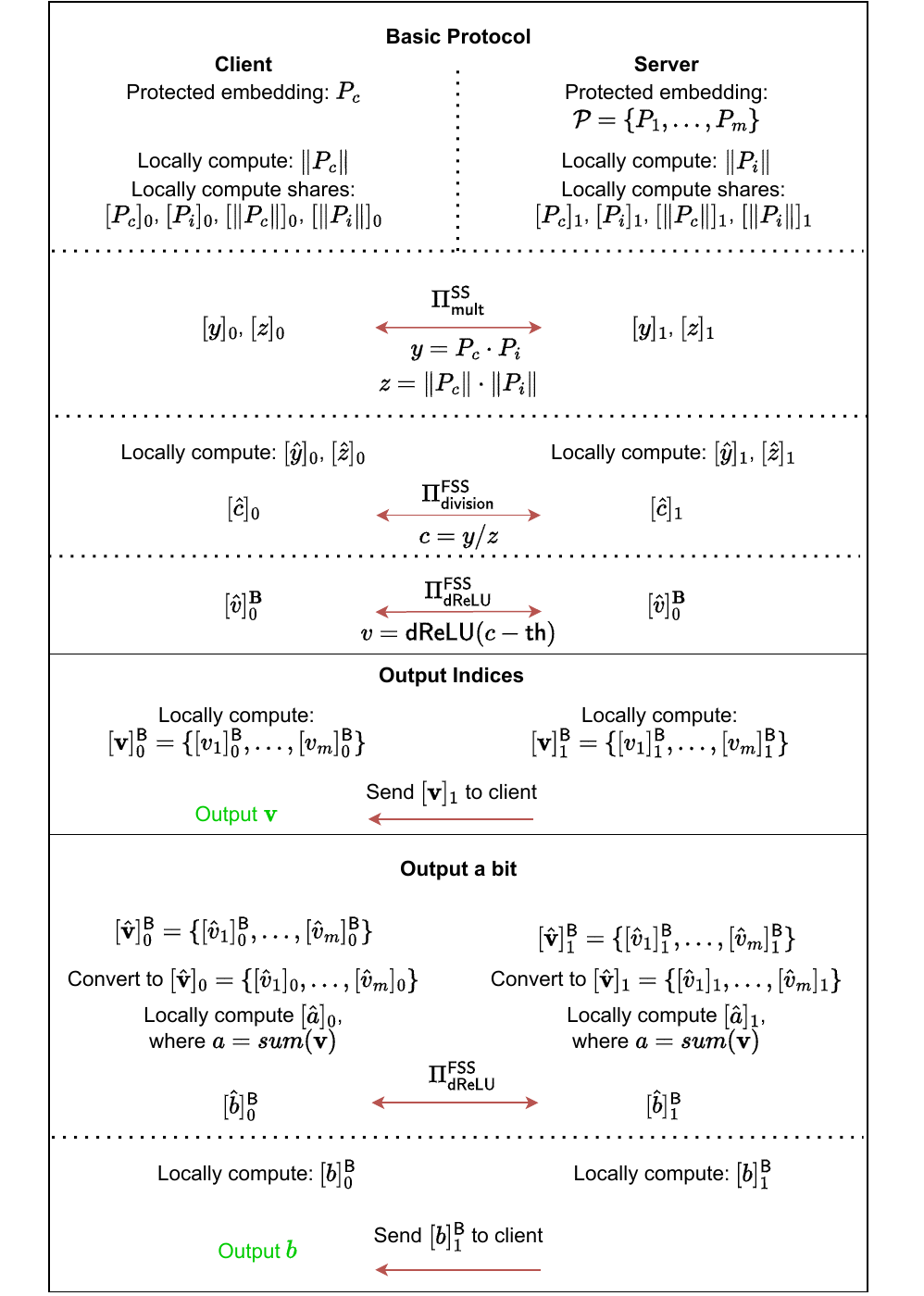}
    \caption{Face embedding distance comparison protocol $\Pi^\mathsf{FSS}_\mathsf{EmComp}$ via FSS}
    \label{fig:fss-protocol}
\end{figure}

To solve this issue, we propose an FSS-based face embedding distance comparison protocol.
During the protocol execution, sensitive intermediate results (e.g. cosine distance) remain hidden from both parties.
We demonstrate our protocol $\Pi^\mathsf{FSS}_\mathsf{EmComp}$ in Figure \ref{fig:fss-protocol}.

%\subsubsection{Basic Protocol}
\textbf{Basic Protocol.}
Within the basic protocol part of $\Pi^\mathsf{FSS}_\mathsf{EmComp}$, parties securely compute the cosine distance for each face embedding pair then compare it to a precomputed threshold value $\mathsf{th}$.
For clarity, we focus on a single embedding pair comparison in the following description. Since the client holds $P_c$ and the server holds $P_i$ as private inputs, they can locally compute the Euclidean 2-norm of $P_c$ and $P_i$, separately. 
Then parties secretly share their local results including $P_c$, $\| P_c \|$, $P_i$, $\| P_i\|$ by locally generating pseudo-randomness from a pseudo-randomness function (PRF).
We note the shared value as $[\cdot]$.
After executing the multiplicative protocol (two times in a single round), parties receive $[y]$ and $[z]$, where $[c] = [y]/[z]$ outputs the desired cosine distance result.
Before executing the FSS-based division protocol $\Pi^\mathsf{FSS}_\mathsf{division}$, parties use a pre-defined randomness to mask both $[y]$, $[z]$ to $[\hat{y}]$, $[\hat{z}]$.\footnote{We denote the masked value as $\hat{\cdot}$.}
Using the computed result $[\hat{c}]$ of $\Pi^\mathsf{FSS}_\mathsf{division}$, parties compute $[\hat{c}] - \mathsf{th}$, and execute the protocol $\Pi^\mathsf{FSS}_\mathsf{dReLU}$ to obtain the (masked) comparison result $[\hat{v}]^\mathsf{B}$.\footnote{We use $[\cdot]^\mathsf{B}$ to represent a boolean shared value}
If $v = 0$, the cosine distance of the compared embedding pair is smaller than the threshold value $\mathsf{th}$.

%\subsubsection{Matched Embedding Indices vs. Single Bit Output}
\textbf{Matched Embedding Indices vs. Single Bit Output.}
In the first application scenario, we assume that the client can directly receive the comparison result $\mathbf{v} = \{v_1,...,v_m\}$.
Thus, parties can simply convert the masked comparison result $[\hat{\mathbf{v}}]$ back to $[\mathbf{v}]$ then reveal it to the client.
The protocol ends here.

In some other applications (e.g. an authentication application), the client should only be aware of whether there is a matched embedding existing in the database but nothing else.
As a result, directly revealing the comparison result $\mathbf{v} = \{v_1,...,v_m\}$ to the client leaks additional information (e.g. the number of matched embeddings).
For this scenario, we construct the third part of our protocol as shown in Figure~\ref{fig:fss-protocol}.
After computing $[\hat{\mathbf{v}}]^\mathsf{B}$, parties execute a conversion protocol to convert the boolean shared $[\hat{\mathbf{v}}]^\mathsf{B}$ to arithmetically shared $[\hat{\mathbf{v}}]$.
Then parties simply compute the sum of all entities in $[\hat{\mathbf{v}}]$, which is denoted as $[\hat{a}]$.
We notice that if $a > 0$, there is at least one matched embedding stored in the database, which is exactly the output that the client is supposed to obtain.
After executing the comparison protocol $\Pi^\mathsf{FSS}_\mathsf{dReLU}$ (also known as the dReLU protocol), parties simply convert the result back to $[b]^\mathsf{B}$ and reveal it to the client.

% An FSS-based solution differs from SS-based approaches mainly in the dReLU (comparison) protocol. We adopt the SS-based protocol implemented in \cite{dai2023force}, 
% where the comparison computation is performed by letting parties compute a prefix parallel adder circuit, 
% which results approximately 5$\ell$ communication overhead across $\mathcal{O}(\mathsf{log}\: \ell)$ rounds in the online stage (we let $\ell$ denote the length of the fixed-point representation). 
% By replacing this computation with $\pi^\mathsf{FSS}_\mathsf{Div}$, the online communication overhead is reduced to $\ell$ across only $\mathcal{O}(1)$ round.

An FSS-based solution differs from SS-based approaches mainly in the dReLU (comparison) protocol. We adopt the SS-based protocol in \cite{dai2023force}, 
where the comparison computation is performed by letting parties compute a prefix parallel adder circuit, 
which results approximately 5$\ell$ communication overhead across $\mathcal{O}(\mathsf{log}\: \ell)$ rounds in the online stage ($\ell$ denote the length of the fixed-point representation). 
By replacing this computation with $\pi^\mathsf{FSS}_\mathsf{Div}$, the online communication overhead is reduced to $\ell$ across only $\mathcal{O}(1)$ round.

% \pi^\mathsf{FSS}_mathsf{Div}%
\section{EXPERIMENTS}\label{sec:pagestyle}

\subsection{Datasets and Implementation Details}

The VGGFace2 \cite{cao2018vggface2} dataset is utilized as the training dataset. For the 1:1 verification, benchmark datasets such as LFW \cite{huang2008labeled}, CFP \cite{sengupta2016frontal} AgeDB \cite{moschoglou2017agedb}, CALFW \cite{zheng2017cross} and CPLFW \cite{zheng2018cross} are used. For the 1:N verification, the dataset is selected from MS-Celeb-1M \cite{guo2016ms} as the scheme in \cite{han2024privacypreserving}. Images have the size in 112 $\times$ 112 pixels. After upsampling and YCbCr color space transformation, images are converted into frequency domain by applying DCT. The DCT implementation is adapted from TorchJPEG \cite{ehrlich2020quantization}. Additionally, the original RGB images are also feed into skin color patch extractor to detect face landmarks. The landmark detection is based on Mediapipe \cite{lugaresi2019mediapipe}. For each image, we select two regions with size 28 $\times$ 28 pixels from both left and right side cheeks and then upsample them 4 times before feature extraction. MobileFaceNet \cite{chen2018mobilefacenets} is used as skin color feature extractor in our work. For dimensionality reduction and information fusion, we use HFCF scheme \cite{han2024privacypreserving} to reduce DCT channel numbers and fuse the frequency information with sparse DLBP (refer to \cite{xiao2020color} for implementation, we assign a value of $10^{-8}$ to $\xi$ to prevent the denominator from being equal to zero.). The implementation of DP model is based on \cite{ji2022privacy}. In order to make a fair comparison with other methods, ResNet34 \cite{He_2016_CVPR} is selected as the FR model backbone and PolyProtect \cite{hahn2022towards} is chosen as the embedding mapping algorithm. For the parameters C, E and overlap in PolyProtect, we use the same as in \cite{han2024privacypreserving}. 
The momentum and weight of the SGD optimizer are set to 0.9 and 0.0005. For the ArcFace loss function, scale and margin are set to 64 and 0.3, respectively. 
The experiments are conducted on two Tesla P100 16G GPUs (NVIDIA).

% Other detailed configurations are demonstrated in Table \ref{tab:model configuration}
% \begin{table}[h]
% \caption{Model training configurations.}\label{tab:model configuration} \centering
% \begin{tabular}{c|c}
%   \hline
%   Para & Value \\
%   \hline
%   GPU & NVIDIA Tesla P100\\
%   Platform & PyTorch\\
%   Optimizer & SGD\\
%   Loss function & ArcFace\\
%   Epoch & 24\\
%   \hline
% \end{tabular}
% \end{table}

\subsection{Performance Comparisons}

To evaluate the proposed method, we report recognition accuracy in two different settings including 1:1 face verification and 1:N face verification. 1:1 means the determination of whether a given pair of faces is matched or not, while in the 1:N case, the test face is compared with a set of faces in the database to find the most likely match. ArcFace \cite{deng2019arcface} is served as RGB image baseline; DCTDP \cite{ji2022privacy} is regarded as privacy-preserving noised DCT input baseline; HCFC-DLBP \cite{han2024privacypreserving} is the fused frequency and color input baseline. It is worth to notice that they use the combination of DCT and DLBP for FR task, however, there is no previous experiment about whether only using DLBP is possible for recognition. Thus, we also train DLBPDP model which takes the noised DLBP feature as input for testing the performance of DLBP itself. 

The accuracy comparison is shown in Table \ref{tab:1:1 verification} and Table \ref{tab:1:N verification} \footnote{ArcFace is FR benchmark without privacy-preserving.}. 
DLBPDP can reach certain performance levels compared with DCTDP but with lower accuracy. 
Our method outperforms on 3 out of 5 validation datasets compared with other PPFR approaches in classical 1:1 verification. 
The variation in skin color has minimal impact, as the performance is already very close to 1, therefore, the contribution is minimal.
According to the retrieval rate, our method achieves the highest accuracy in 1:N verification. 
% The improved results indicate pure skin color patches do help for the better recognition.
In this case, the model has more uncertainty and the skin color helps to select the right face more accurately.
The accuracy on protected embeddings is higher than the counterpart on original embedding due to the user-specific mapping property of PolyProtect. For the number of channels, there are 6 extra channels introduced from the left and right face skin color patches in our method. It is crucial to note that the color patch detection may occasionally fail due to the effectiveness of landmark detection and the quality of input images. In order to tackle such situation, the skin color patches are filled with 0 for each pixel. Therefore, for some face images, the skin color feature extractor will only receive pure black patches.

% \begin{table*}[t]
% \caption{1:1 verification accuracy comparison.}\label{tab:1:1 verification} \centering
% \begin{tabular}{c|c|c|c|c|c|c c}
%   \hline
%   Method (\%)  & \# Channels &LFW & CFP-FP  & AgeDB & CALFW & CPLFW\\
%   \hline
%   ArcFace \cite{deng2019arcface}            & 3   & 99.70 & 98.14 & 95.62 & 94.28 & 93.10\\
%   DCTDP \cite{ji2022privacy}                & 189 & 99.64 & 97.69 & \textbf{95.10} & \textbf{93.87} & 91.77\\
%   DLBPDP                                     & 64 & 99.62 & 96.93 & 93.77 & 92.77 & 91.78\\
%   HFCF-DLBP \cite{han2024privacypreserving} & 126 & 99.57 & 97.69 & 95.03 & 92.95 & 91.70\\
%   Ours                                      & 132 & \textbf{99.67} & \textbf{97.71} & 94.91 & 93.58 & \textbf{91.80}\\
%   \hline
% \end{tabular}
% \end{table*}

\begin{table*}[t]
  \caption{1:1 verification accuracy comparison.}\label{tab:1:1 verification} \centering
  \begin{tabular}{c|c|c|c|c|c|c|c c}
    \hline
    Method (\%)  & Privacy-Preserving         & \# Channels & LFW & CFP-FP  & AgeDB & CALFW & CPLFW\\
    \hline
    ArcFace \cite{deng2019arcface}            & No    & 3   & 99.70 & 98.14 & 95.62 & 94.28 & 93.10\\\cline{1-8}
    DCTDP \cite{ji2022privacy}                & Yes   & 189 & 99.64 & 97.69 & \textbf{95.10} & \textbf{93.87} & 91.77\\
    DLBPDP                                    & Yes   & 64  & 99.62 & 96.93 & 93.77 & 92.77 & 91.78\\
    HFCF-DLBP \cite{han2024privacypreserving} & Yes   & 126 & 99.57 & 97.69 & 95.03 & 92.95 & 91.70\\
    Ours                                      & Yes   & 132 & \textbf{99.67} & \textbf{97.71} & 94.91 & 93.58 & \textbf{91.80}\\
    \hline
  \end{tabular}
  \end{table*}

% \begin{table*}[t]
% \caption{1:N verification accuracy comparison on original embeddings and protected embeddings (based on PolyProtect with overlap 4).}\label{tab:1:N verification} \centering
% \begin{tabular}{c|c|c|c|c|c|c|c}\hline
%       Method (\%) & \# Channels & \multicolumn{6}{c}{Retrieval Rate $\uparrow$}\\\cline{3-8}
%        & & \multicolumn{3}{c}{Original Embedding} & \multicolumn{3}{|c}{Protected Embedding}\\\cline{3-8}
%        & & Rank=1 & Rank=5 & Rank=10 & Rank=1 & Rank=5 & Rank=10\\\cline{1-8}
%       ArcFace \cite{deng2019arcface} & 3 & 87.8 & 93.8 & 95.3 & 95.2 & 97.3 & 98.1\\\hline
%       DCTDP \cite{ji2022privacy} & 189 & 79.3 & 86 & 88.3 & 88.7 & 93.9 & 95.3\\\hline
%       DLBPDP & 64 & 78.3 & 84.9 & 87.8 & 87.2 & 93.7 & 95.6\\\hline
%       HFCF-DLBP \cite{han2024privacypreserving} & 126 & 81.9 & 90.3 & 92.5 & 91.2 & 95.7 & 96.8\\\hline
%       Ours & 132 & \textbf{83.1} & \textbf{91.0} & \textbf{93.0} & 91.7 & 95.7 & 96.7\\\hline
% \end{tabular}
% \end{table*}

\begin{table*}[t]
  \caption{1:N verification accuracy comparison on original embeddings and protected embeddings (based on PolyProtect with overlap 4).}\label{tab:1:N verification} \centering
  \begin{tabular}{c|c|c|c|c|c|c|c|c}\hline
        Method (\%) & Privacy-Preserving & \# Channels & \multicolumn{6}{c}{Retrieval Rate $\uparrow$}\\\cline{3-9}
         & & & \multicolumn{3}{c}{Original Embedding} & \multicolumn{3}{|c}{Protected Embedding}\\\cline{4-9}
         & & & Rank=1 & Rank=5 & Rank=10 & Rank=1 & Rank=5 & Rank=10\\\cline{1-9}
        ArcFace \cite{deng2019arcface} & No & 3 & 87.8 & 93.8 & 95.3 & 95.2 & 97.3 & 98.1\\\hline
        DCTDP \cite{ji2022privacy} & Yes & 189 & 79.3 & 86 & 88.3 & 88.7 & 93.9 & 95.3\\
        DLBPDP & Yes & 64 & 78.3 & 84.9 & 87.8 & 87.2 & 93.7 & 95.6\\
        HFCF-DLBP \cite{han2024privacypreserving} & Yes & 126 & 81.9 & 90.3 & 92.5 & 91.2 & \textbf{95.7} & \textbf{96.8}\\
        Ours & Yes & 132 & \textbf{83.1} & \textbf{91.0} & \textbf{93.0} & \textbf{91.7} & \textbf{95.7} & 96.7 \\\hline
  \end{tabular}
  \end{table*}

\subsection{Black-box Attacking}

This section addresses the trustworthiness of privacy protection in the proposed HFCF-SkinColor PPFR model through the analysis of black-box attacking. We assume attacker can obtain outputs after all the privacy-preserving preprocessing. The objective of attacking experiments is to reconstruct the original face image. Nevertheless, attackers have the capability to gather large quantities of face images from websites or any publicly available face datasets. In the most challenging scenario, we assume attackers can gain access to the set of face images that were utilized to train the PPFR model. In this case, we randomly select 15690 images from our training dataset. The processed inputs are used to instruct a decoder to reconstruct original face images. Ultimately, attackers can take advantage of a trained decoder to retrieve the user's face image. We employ UNet \cite{ronneberger2015u} as the decoder for reconstructing original images from processed images. During the training phase, we use the SGD optimizer with a learning rate of 0.0001 for 10 epochs and batch size 64. 

\begin{figure}[hbt!]
\centering
\includegraphics[scale=0.33]{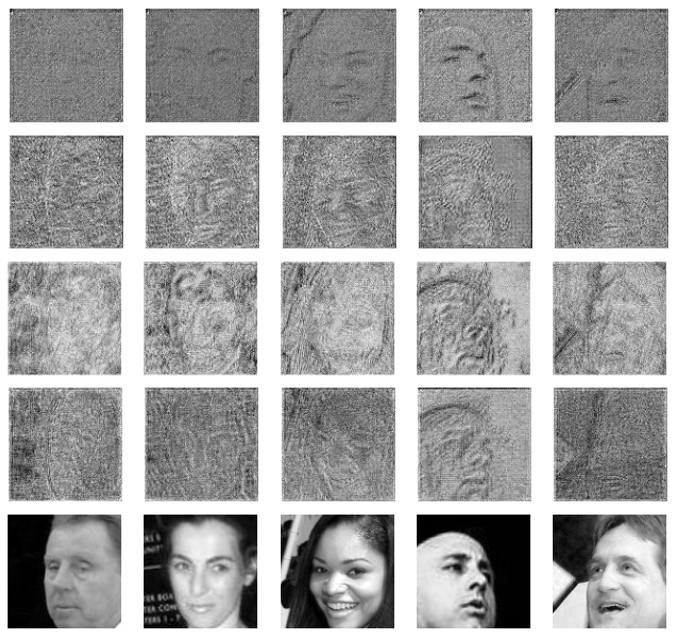}
\caption{Face reconstruction by black-box attacking. From top to bottom, the target model are DCTDP, DLBPDP, HFCF-DLBP and Ours. The last row are ground truth images shown in gray color for better visualization.}
\label{fig:blackbox attack}
\end{figure}

\begin{figure}[hbt!]
\centering
\includegraphics[scale=0.4]{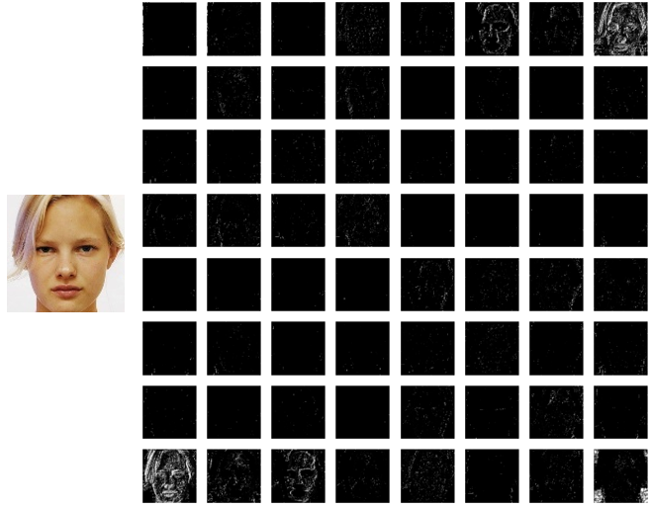}
\caption{Test image and DLBP representation.}
\label{fig: DLBP}
\end{figure}

In Figure \ref{fig:blackbox attack}, the reconstructed face images are presented from 5 identities. The naive UNet model can easily recovery main face structures like eye browns, eyes, nose and lip in case of DCTDP model even though DP noise is added during training. Results from DLBPDP and HFCF-DLBP are much noisier compared with DCTDP since the face attributes representation from DLBP is more sparse (as mentioned in \cite{han2024privacypreserving}, only very few of DLBP images contain visual information. The example DLBP visualization is in Figure \ref{fig: DLBP}).   

\begin{figure}[h!]
\centering
\includegraphics[scale=0.5]{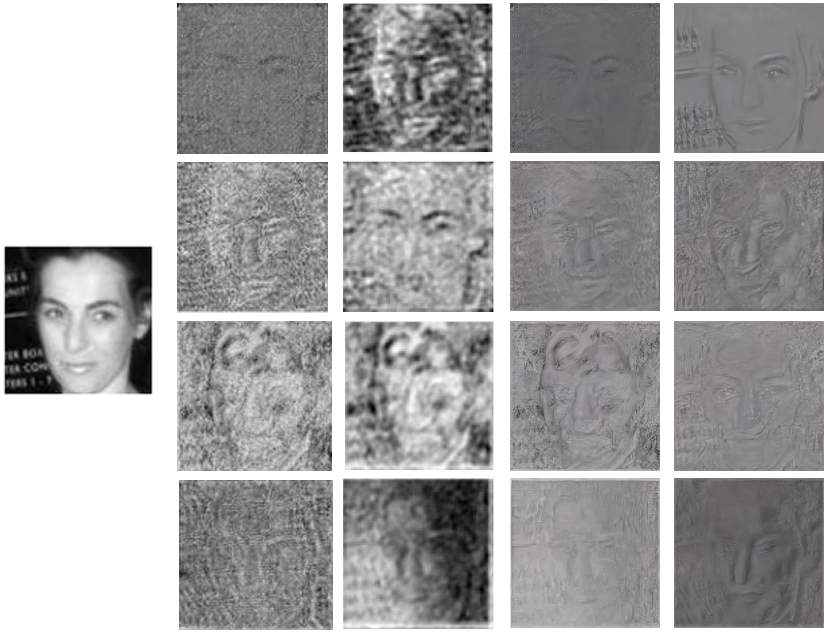}
\caption{Image restoration on a UNet reconstructed image. From top to bottom, target models are DCTDP, DLBPDP, HFCF-DLBP and Ours. The test image, results from UNet, results after applying the wiener filter and GFP-GAN are shown from left to right. The most right column contains the results from GFP-GAN when UNet reconstruct image without adding DP noise.}
\label{fig: wiener filter and GFPGAN}
\end{figure}

The reconstructed face images from ours are similar to results from HFCF-DLBP without obvious improvement in terms of clearness of facial structure. This observation indicates that including the skin color patch does not improve the face reconstruction attack.
% \ricardo{that including the skin color patch does not improve the face reconstruction attack.}
%even with skin color patch revealing, it is difficult to utilize such information for face reconstruction attack.
As the direct reconstructed image from naive UNet might suffer noise artifacts, the classical image denoising method wiener filter \cite{anderes2012robust} and the well-known generative adversarial network (GAN) based face restoration model GFP-GAN \cite{wang2021towards} (CVPR 2021) are applied to further restore the face images. The restored images are showing in Figure \ref{fig: wiener filter and GFPGAN}.

The wiener filter method is able to highlight the facial structure from UNet result, however, the face area still contains much noise in comparison with GFP-GAN approach. In the black-box attacking scenario, the naive UNet is trained based on full processed inputs (with DP noise added). In the extreme case, if DP model is bypassed somehow, it is easy to reconstruct clear face. Especially for frequency-based model DCTDP, the background of original face image can be also restored by GFP-GAN (refer the last column in Figure \ref{fig: wiener filter and GFPGAN}).

\subsection{Benchmark FSS}

\begin{table}
    \centering
    \caption{1:N face embedding comparison protocol benchmarks (millisecond) in LAN and WAN.}
    \begin{tabular}{c c | r r r }
        \hline
        Network & Protocol  & Basic & Indices & Bit\\
        \hline
        \multirow{2}{*}{LAN} & SS & 9670 & \textbf{0.45} & 5.74 \\
        %\hline
        & FSS & \textbf{5610} & 1.26 & \textbf{1.63} \\
        \hline
        \multirow{2}{*}{WAN} & SS & 18667 & \textbf{17.67} & 544.67 \\
        %\hline
        & FSS & \textbf{10833} & 20.12 & \textbf{43.43} \\
        \hline
    \end{tabular}
    \label{tab:benchmark}
\end{table}

Experiments ran on a server with 2 CPUs, \texttt{Intel(R) Xeon(R) Platinum 8360Y CPU @ 2.40GHz}, and \texttt{16 $\times$ 128GB} of RAM.
We set the number of threads to 12, and we run each experiment 10 times and take the average running time in the online stage.
For network settings, we use \texttt{tc} tool~\footnote{https://man7.org/linux/man-pages/man8/tc.8.html} to simulate both LAN with 1\texttt{Gbps} bandwidth + 0.2\texttt{ms} round-trip latency and WAN with 100\texttt{Mbps} bandwidth + 40\texttt{ms} round-trip latency. To implement the secret sharing protocols we used the library from \cite{dai2023force}. Since our pipeline is in python, we created python bindings for the c++ functions using pybind11~\cite{pybind11}, which add negligible overhead.

We benchmark each part of the protocol $\Pi^\mathsf{FSS}_\mathsf{EmComp}$ in Table~\ref{tab:benchmark}.
As a result, the FSS-based protocol dominates the SS-based protocol in both LAN and WAN settings except for the \textbf{Indices Output Protocol}, which introduces only minor magnitude.
We observe that the FSS-based protocol improves the running time of the basic protocol by about 42\%.
And specifically, it reduces the running time of the \textbf{Bit Output Protocol} by about 92\% in WAN setting.
Such improvement is introduced by the reduced communication rounds required in an FSS-based solution within the dReLU protocol (significant for a single value comparison).

\section{Conclusion}

In this work we present a skin color patch extraction as the auxiliary module for improving existing Privacy-Preserving Face Recognition methods without privacy deterioration. The usage of pure skin color information can improve overall recognition accuracy.
In addition, we propose a Functional Secret Sharing-based face embedding comparison protocol to securely and efficiently compare protected embeddings.
The potential limitation of our method is that the face skin color can be revealed if such auxiliary module is deployed at third-party platform. 
One potential solution is to integrate this module into a secure isolation hardware such as Trusted Execution Environment.
In the future work, we may reduce the complexity of feature extractor for more efficient training.
% TrustZone

% References should be produced using the bibtex program from suitable
% BiBTeX files (here: strings, refs, manuals). The IEEEbib.bst bibliography
% style file from IEEE produces unsorted bibliography list.
% -------------------------------------------------------------------------
\bibliographystyle{IEEEbib}
\bibliography{ref,security}

\end{document}